\pdfoutput=1

\documentclass[11pt]{article}

\usepackage{EMNLP2023}

\usepackage{times}
\usepackage{latexsym}

\usepackage[T1]{fontenc}

\usepackage[utf8]{inputenc}

\usepackage{microtype}

\usepackage{inconsolata}

\usepackage{booktabs}       
\usepackage{multirow}
\usepackage{threeparttable}
\usepackage{graphicx}
\usepackage{subfigure}
\usepackage{xspace}
\usepackage{amssymb} 

\def\figref#1{Figure~\ref{fig:#1}}
\def\figlabel#1{\label{fig:#1}\label{p:#1}}

\def\tabref#1{Table~\ref{tab:#1}}

\def\tablabel#1{\label{tab:#1}\label{p:#1}}

\def\secref#1{\S\ref{sec:#1}}
\def\seclabel#1{\label{sec:#1}}

\def\method{CoThought\xspace}

\newcommand\blfootnote[1]{%
  \begingroup
  \renewcommand\thefootnote{}\footnote{#1}%
  \addtocounter{footnote}{-1}%
  \endgroup
}

\title{Baby's CoThought: Leveraging Large Language Models for Enhanced Reasoning in Compact Models}

\vspace{5pt}
\author{
Zheyu Zhang$^{\ast}$$^{\clubsuit}$~~
Han Yang$^{\ast}$$^{\clubsuit,\diamondsuit}$~~
Bolei Ma$^{\ast}$$^{\spadesuit}$~~
David Rügamer$^{\spadesuit,\heartsuit}$~~\smallskip\vspace{3pt}
Ercong Nie$^{\dag}$$^{\clubsuit,\heartsuit}$\\
$^\clubsuit$Center for Information and Language Processing, LMU Munich\\
$^\diamondsuit$GESIS - Leibniz Institute for the Social Sciences, Cologne\\
$^\spadesuit$Department of Statistics, LMU Munich ~~~\smallskip\vspace{3pt}
$^\heartsuit$Munich Center for Machine Learning \\
\texttt{zheyu.zhang@campus.lmu.de} ~~~~~~ \texttt{han.yang@gesis.org}\\
\texttt{\{bolei.ma, david.ruegamer\}@stat.uni-muenchen.de}\\
\texttt{nie@cis.lmu.de}
}

\begin{document}
\maketitle

\begin{abstract}
Large Language Models (LLMs) demonstrate remarkable performance on a variety of natural language understanding (NLU) tasks, primarily due to their in-context learning ability. 
This ability could be applied to building baby-like models, i.e. models at small scales, improving training efficiency. 
In this paper, we propose a ``\textbf{CoThought}'' pipeline, which efficiently trains smaller ``baby'' language models (BabyLMs) by leveraging the \textbf{C}hain \textbf{o}f \textbf{Thought} 
prompting of LLMs. Our pipeline restructures a dataset of less than 100M in size using GPT-3.5-turbo, transforming it into task-oriented, human-readable texts that are comparable to the school texts for language learners. The BabyLM is then pretrained on this restructured dataset in a RoBERTa fashion. In evaluations across 4 benchmarks, our BabyLM outperforms the vanilla RoBERTa in 10 linguistic, NLU, and question-answering tasks by more than 3 points, showing a superior ability to extract contextual information.\blfootnote{$^\ast$ Equal contribution.}\blfootnote{$^\dag$ Corresponding author.} These results suggest that compact LMs pretrained on small, LLM-restructured data can better understand tasks and achieve improved performance.\footnote{The code for data processing and model training is available at: \url{https://github.com/oooranz/Baby-CoThought}.}
\end{abstract}

\section{Introduction}
Recent advances in language modeling of Large Language Models (LLMs) have shown great performance potential on diverse NLP tasks. 
A large number of work has been proposed towards enhancing LLMs pretraining at massive scales 
\citep{devlin-etal-2019-bert,gpt,gpt3}.
However, less attention has been paid to language model (LM) pretraining at smaller human-like data scales, i.e.\ smaller data scales, which are similar to the amount of language data for human language acquisition.


Studies in language acquisition demonstrate that humans predominantly acquire language in early life stages by observing their environment. Significant progress in language communication and usage is typically achieved by early childhood \citep{tomasello2003, saxton2010}. 
Previous studies show that language modeling is to some extent similar to children's language acquisition, 
as they both require input data from the outside world and learn the data by updating knowledge about the outside world repeatedly \citep{nikolaus-fourtassi-2021-modeling, chang-bergen-2022-word, evanson-etal-2023-language}.
It is reasonable to apply this human cognitive process to LM pretraining by using relatively small sets of pretraining data that are comparable to the text data for human language acquisition.

While a child learns a piece of knowledge by continuously obtaining relevant examples from the outside world and updating its knowledge base, pretrained LLMs have the capacity to learn and complete previously unknown tasks when given several task samples or instructions already from the inside of their context, 
the process of which is known as ``\textit{In-Context Learning}'' (ICL) \citep{gpt3}. A more recent advance of ICL called ``\textit{Chain of Thought}'' (CoT) \citep{wei2022chain} significantly enhances the reasoning abilities of LLMs. CoT enables LLMs to perform a series of intermediate reasoning steps by providing a few CoT demonstrations as examples during the training process. This method has been found to be very effective, especially in complex reasoning tasks.


\begin{figure*} [ht]
\centering
\includegraphics[scale=0.75]{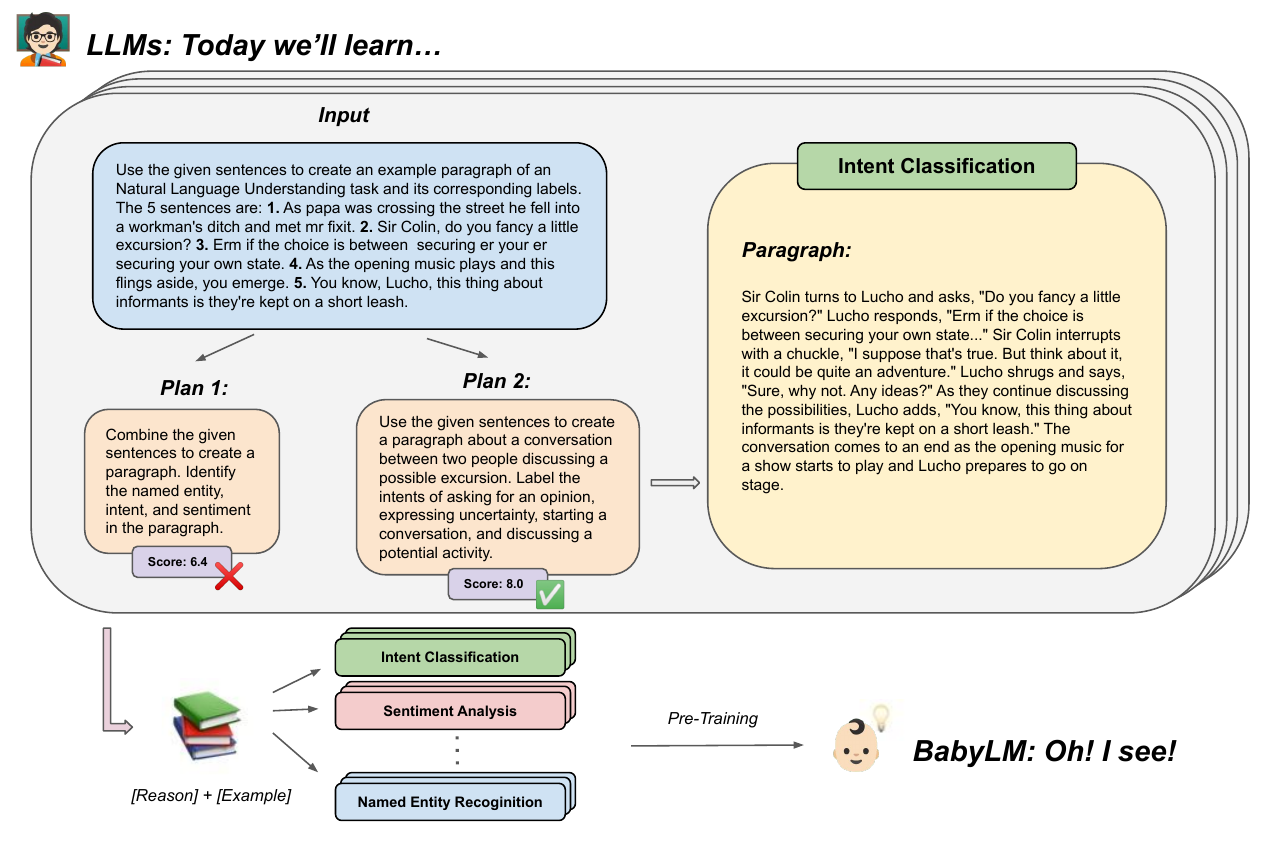}
\caption{Overview of the ``\method'' pipeline. We propose to generate NLU examples from discrete short sentences using CoT prompting and an automatic scoring mechanism. This constructs a pretraining dataset in a \textit{[Reason] + [Example]} format, which is then used to pretrain smaller models.}
\figlabel{cothought}
\end{figure*}

The LLM is like a teacher who is able to transfer knowledge by 
reformulating raw data from the outside world into a task-like text format by CoT prompting, making the data more suitable for teaching.
The BabyLM is like a student who is trained based on this generated text. In this work
, we propose ``\textbf{\method}'' pipeline to pretrain a BabyLM with human-like smaller corpus data, by leveraging the LLM's \textbf{C}hain \textbf{o}f \textbf{Thought} feature and the child's cognitive learning ability. In this way, the LLM and the child are ``co-thinking'' during the training process. 
We use the ``\method'' approach to train our BabyLM, combining the productivity of the LLM with the effectiveness of human language acquisition for LM pretraining.


Our overall framework is illustrated in \figref{cothought}. The raw pretraining data is provided by \citet{warstadt-et-al-2023-babylm} in the BabyLM Challenge, which has the goal of sample-efficient pretraining on a developmentally plausible corpus at a small human-like data scale. We choose the loose track of the BabyLM Challenge, where we apply our ``\method'' pipeline and use the LLM GPT-3.5-turbo\footnote{\url{https://platform.openai.com/docs/models/gpt-3-5}} to preprocess the raw data. 
For every 5 sentences of the raw data, the GPT-3.5-turbo uses CoT prompting to propose different NLU tasks and selects the best task. Then, it combines these 5 sentences into a task-like text based on the best task for our BabyLM to learn. 
The BabyLM is pretrained on the augmented data in a RoBERTa \citep{roberta} fashion. 
Our BabyLM pretrained in the \method pipeline notably
outperforms the original RoBERTa model on common benchmarks.

Our work makes contributions in 
\begin{itemize}
    \item[1)] proposing the \method pretraining pipeline fitting the human-like data scenarios,
    \item[2)] pretraining a BabyLM model of the RoBERTa-base architect in the \method pipeline surpassing the original RoBERTa model on several tasks, and
    \item[3)] providing insights of the \method pipeline by conducting linguistic case analysis on representative tasks.
\end{itemize}

\section{Related Work}

\paragraph{Language Acquisition and Modelling} The language acquisition of children is a widely studied topic in linguistics. The empiricism of language acquisition contends that language ability is a component of social cognitive ability and children acquire language through language communication and language use \citep{bybee2001, pullum2002, tomasello2003, saxton2010}. 
According to the Universal Grammar \citep{chomsky1957}, language norms and parameters are hard-wired within every single person, and learning a language is just a matter of adjusting those parameters \citep{gegov2014modelling}. 
In this way, child language acquisition and language modeling are similar, as the neural language models such as BERT \citep{devlin-etal-2019-bert} and GPT \citep{gpt} are pretrained based on big corpora with their model parameters tuned during pretaining. Recent studies show the applicability of language models to child language development tracking. \citet{nikolaus-fourtassi-2021-modeling} propose an integrated perception- and production-based learning and highlight that children are not only understood as passively absorbing the input but also as actively participating in the construction of their linguistic knowledge in language learning. \citet{chang-bergen-2022-word} study the factors that predict words’ ages of acquisition in contemporary language models compared to word acquisition in children. \citet{evanson-etal-2023-language} compare the sequence of learning stages of language models with child language acquisition.

\paragraph{In-Context Learning (ICL)} LLMs like GPT-3 \citep{gpt3} make ``\textit{In-Context Learning}'' possible, which means the model makes predictions by learning from a natural language prompt describing the language task or learning from (only a few) examples. Based on the concept of ICL, recent research has demonstrated that LLMs can be used to extract relevant knowledge from the content. \citet{liu-etal-2022-generated} propose to use GPT-3 to generate pertinent contexts and then supply those contexts as extra input in order to answer a commonsense question. \citet{yu2023generate} employ a generate-then-read pipeline which first prompts a large language model to generate contextual documents based on a given question, and then reads the generated documents to produce the final answer.

\paragraph{Chain of Thought (CoT)} \citet{wei2022chain} introduced ``\textit{Chain of Thought}'', which is a series of intermediate reasoning steps a few chain of thought demonstrations are provided as exemplars in prompting, in order to improve the ICL ability of LLMs to perform complex reasoning. \citet{kojima2023large} demonstrate the zero-shot performance of CoT. \citet{paranjape2023art} introduces a framework that uses frozen LLMs to automatically generate intermediate reasoning steps as a program. \citet{yao2023tree} put forward a “\textit{Tree of Thoughts}” (ToT) framework, which generalizes over CoT to prompting language models and enables exploration over coherent units of text (“thoughts”) that serve as intermediate steps toward problem solving. A more recent study \citep{gu-etal-2023-pre} proposes a pretraining for ICL framework which pretrains the model on a set of “intrinsic tasks” in the general plain-text corpus using the simple language modeling objective to enhance the language models’ ICL ability. 


\section{Method}

In the realm of cognitive learning, the teacher's thought process greatly influences the way instructional content is delivered, which in turn impacts the students' understanding \citep{chew2021cognitive}. Our method attempts to mimic this process. The LLMs, in the role of the teacher, use CoT prompting to reinterpret the raw data, generating task-like text that incorporates the context of the sentences and enriches the learning materials.

We first introduce an overview of our CoThought pipeline (see \figref{cothought} for an illustration) and then describe the details in the following sections. 

\subsection{Problem Statement}
The genesis of our research lies in addressing a significant problem within the context of the BabyLM Challenge as proposed by \citet{warstadt-et-al-2023-babylm}. The goal of this challenge is to conduct sample-efficient pretraining on a developmentally plausible corpus at a small human-like data scale, which we previously introduced. Nevertheless, the majority of the training data provided consists of discrete short sentences. As an illustration, below are some of the provided sentences:

\begin{itemize}
\small
\item[-] \texttt{You want your book back, don't you?}
\item[-] \texttt{Let's see, do you want to see who this is?}
\item[-] \texttt{This is Big Bird.}
\item[-] \texttt{Enough with that.}
\item[-] \texttt{Can you read your book again? You like the book?}
\end{itemize}

These sentences, albeit contextually rich, are sampled from a wide range of sources including dialogues, scripted content, fiction, nonfiction, and child-directed materials. Due to the diverse and fragmented nature of this dataset, the sentences often lack strong semantic ties with each other, making it difficult for models to learn contextual and coherent representations.

In response, we propose a method that transforms these fragmented sentences into cohesive units using LLMs, subsequently enabling more effective learning for the smaller models. The succeeding sections will provide a succinct outline of our pipeline and process.

\subsection{Creative NLU-Example Generation}
Inspired by recent studies that demonstrate the capability of LLMs to generate rationales supporting their predictions, we invent a novel task called Creative NLU-Example Generation (CNLU-EG), inspired by the Creative Writing task proposed by the ``\textit{Tree of Thought}'' \citep{yao2023tree}. Instead of creating coherent paragraphs from random sentences, CNLU-EG employs the provided sentences to generate coherent paragraphs, which define a plausible intrinsic NLU task and its corresponding labels. In this task, we employ the reasoning capability of LLMs to generate rationales for training smaller baby models.

We first remove any duplicate sentences from the BabyLM\_100M \citep{warstadt-et-al-2023-babylm} \(D\). After the cleaning process, we randomly sample five unique sentences \(\{x_i\}_{i\in D}\) from the cleaned dataset \(D\). We initiate the task by providing a specific CoT prompt $p$ to the LLM. This prompt instructs the LLM to first create a plan, then use the provided sentences to compose an example paragraph that illustrates a possible intrinsic NLU task, and finally generate the corresponding labels for this task. Given the creative nature of the task, we use a zero-shot prompt here. 
The prompt is structured such that it encourages the LLM to present the output in four distinct sections: the plan, the paragraph, the task, and the labels. 

Once the LLM receives the prompt $p$, for each sentence $x_i, i \in D$, the LLM generates an execution plan $\hat{r}_i$, a paragraph $\hat{e}_i$ embodying an example of a possible NLU task, the task name $\hat{t}_i$, and the corresponding labels $\hat{y}_i$.

CNLU-EG essentially transforms the original, discrete sentences into a structured task, anchoring the sentences to a common theme or question. This `taskification' process helps to create a more cohesive narrative, enabling the baby model to gain a more contextual and comprehensive understanding of the sentences.

We also incorporate a scoring mechanism, to assess the coherence of the generated content. We use a separate simple zero-shot prompt, $p_s$, to instruct the LLM to analyze the composed paragraph and assign a coherence score ranging from 1 to 10. For each task output, the LLM generates five such coherence scores from the same scoring prompt $p_s$, and these scores are then averaged to produce a final coherence score. According to our settings, we explicitly direct the LLM to generate two distinct plans for each task. Each plan is independently scored, and the one that achieves a higher coherence score is selected for subsequent steps. 

In this way, the LLM functions as a teacher, generating examples of possible NLU tasks, providing insights into how these examples were created, and supplying the corresponding labels. This collection of generated plans and example paragraphs forms the training data for the smaller model to learn from.

\subsection{Training Data Construction}
Our objective is to construct a high-quality dataset for pretraining our small model, ensuring the instances included in the training set are coherent and task-relevant. As previously discussed, each instance in our data comprises a tuple: an example $e$ and a corresponding plan $r$, denoted as $[e, r]$. However, not all generated instances meet the quality criteria necessary for effective learning.

To filter out lower-quality instances, we employ the coherency score obtained through the $p_s$ prompt. We set a threshold, stipulating that only instances with a coherency score of $s \geq 7.0$ are included in the training data. This threshold was empirically established based on extensive manual analysis to ensure a satisfactory level of coherence and quality in the dataset. Mathematically, this can be represented as:

\begin{equation}
D_{select} = {[e_i, r_i] : i \in D, s_i \geq 7.0}
\end{equation}
Here, $D$ denotes the initial set of generated instances and $D_{select}$ represents the selected high-quality instances that are used for training.

Another important aspect of our methodology is leveraging the correlation between segments with similar intrinsic tasks. Studies indicate that such segments when grouped together, provide valuable information for ICL \citep{gu-etal-2023-pre}. Therefore, we aim to collate instances with similar tasks, denoted as $T$, into grouped sets, which we denote as $G_T$.

\begin{equation}
G_T = {[e_i, r_i] : i \in D_{select}, t_i = T}
\end{equation}

In the equation above, $t_i$ represents the task type of the $i$-th instance, and $G_T$ denotes the set of instances from $D_{select}$ that are associated with task type $T$.

In the end, we amalgamate these grouped sets to create a comprehensive pretraining dataset containing $N$ instances.

\begin{equation}
D_{pretrain} = \bigcup_{T \in \mathcal{T}} G_T
\end{equation}
Here, $\mathcal{T}$ represents the set of all task types and $G_T$ denotes the set of instances corresponding to each task type $T$ in $D_{select}$.

Through these rigorous steps, we ensure that the final training data is both high-quality and task-relevant, optimally structured to facilitate effective learning in our small model.


\section{Experimental Setups}

We conducted our experiments in three parts, the generation of the additional data used for training, the pretraining of the language model, and the evaluation. 

\subsection{Data Generation via CoT Prompting}

We generated first our extended data based on the dataset \texttt{babylm\_100M} \citep{warstadt-et-al-2023-babylm}, which contains subsets including AOCHILDES, BNC spoken, cbt, children stories, Gutenberg, pen subtitles, qed, simple Wikipedia, switchboard, and Wikipedia.\footnote{The full datasets could be downloaded here: \url{https://github.com/babylm/babylm.github.io/raw/main/babylm_data.zip}}

We leveraged the API of GPT-3.5-turbo from OpenAI and provided CoT prompt with the format:

\begin{itemize}
\small
\item[-] \texttt{Use the given sentences to create an example paragraph of an NLU task and its corresponding labels. The 5 sentences are: {input}.}
\item[-] \texttt{Make a plan then write and determine. Your output should be of the following format:}
\item[-] \texttt{Plan:}
\item[-] \texttt{Your plan here.}
\item[-] \texttt{Paragraph:}
\item[-] \texttt{Your paragraph here.}
\item[-] \texttt{Task:}
\item[-] \texttt{[Only the task name here, without additional information.]}
\item[-] \texttt{Labels:}
\item[-] \texttt{[Only the labels here, without additional information.]}
\end{itemize}

The GPT will generate the corresponding answers in the defined format. To evaluate the generated task plans, we prompt the GPT again with the score prompt in the format:

\begin{itemize}
\small
\item[-] \texttt{Analyze the following paragraph, then at the last line conclude ``Thus the coherency score is {s}'', where s is an integer from 1 to 10.}
\end{itemize}

We filter out the generated texts with a score lower than 7. The additional data will be generated by the GPT with the selected proposals as prompts.  

\subsection{Pretraining}

We then trained a RoBERTa model with the extended dataset using \texttt{RobertaForMaskedLM} provided by the \texttt{huggingface} library \footnote{\url{https://huggingface.co/docs/transformers/model\_doc/roberta}}, which uses the default settings of \texttt{RobertaConfig} library and is also the same settings as the hyperparameter of the baseline provided by the organizers. In the training phase, we trained 5 epochs using the Trainer provided by the \texttt{huggingface}. We refer \secref{hyperparameter} for detailed hyperparameters in Appendix. 

\subsection{Benchmarks and Evaluation}
We evaluated the model using the evaluation pipeline tools\footnote{\url{https://github.com/babylm/evaluation-pipeline}}  also provided by the organizer \cite{warstadt-et-al-2023-babylm, eval_harness}. This tool automatically performs experiments on 4 benchmarks: 
\begin{itemize}
    \item[1)] Benchmark of Linguistic Minimal Pairs (BLiMP) \cite{warstadt-etal-2020-blimp-benchmark}; 
    \item[2)] BLiMP Supplement\footnote{The relevant paper for this benchmark had not been published at the time of this project, and the relevant data can be found here \url{https://github.com/babylm/evaluation-pipeline/blob/main/filter\_data.zip}}, including Hypernym,
QA Congruence Easy, QA Congruence Tricky, Subject Aux Inversion, and Turn Taking datasets; 
    \item[3)] General Language Understanding Evaluation (GLUE) \cite{GLUE}, and 
    \item[4)] Mixed Signals Generalization Set (MSGS)  \cite{MSGS}. 
\end{itemize}

The detailed documentation of each benchmark can be found in \secref{full_results}. The organizer \citep{warstadt-et-al-2023-babylm} also provided 3 models as baselines, including OPT-125M, RoBERTa-base, and T5-base, trained on the \texttt{babylm\_100M} data.






\section{Results}
We compare the performance of our BabyLM (trained in the RoBERTa way) to the original RoBERTa-base (baseline). 
\tabref{result-selected} shows our selected experimental results with: i) performance improvement by at least 3 points (+3), and ii) performance reduction over 3 points (-3). 
We report the performance with absolute performance difference of our BabyLM over baseline on the selected tasks, as well as the overall performance of the whole tasks. The full results are available in \secref{full_results}.

\begin{table}[ht]
\centering
\small
\renewcommand\arraystretch{1.4}
\setlength\tabcolsep{8pt}

\begin{tabular}{lccc}
\hline \hline
\multicolumn{1}{c|}{\multirow{2}{*}{\textbf{Tasks}}} & \multicolumn{2}{c|}{\textbf{Models}}                                    & \multirow{2}{*}{\textbf{Diff.}} \\ \cline{2-3}
\multicolumn{1}{c|}{}                                & \multicolumn{1}{c|}{\textbf{Ours}}           & \multicolumn{1}{c|}{\textbf{Baseline}} &                                 \\ \hline
\multicolumn{4}{c}{\textbf{BLiMP}}                                                                                                                               \\ \hline
\multicolumn{1}{l|}{Filler   Gap}                    & \multicolumn{1}{c|}{\textbf{78.52}} & \multicolumn{1}{c|}{68}           & \textbf{10.52}                  \\ \hline
\multicolumn{1}{l|}{Sub.-Verb   Agr.}                & \multicolumn{1}{c|}{\textbf{85.17}} & \multicolumn{1}{c|}{76.2}         & \textbf{8.97}                   \\ \hline
\multicolumn{1}{l|}{Arg.   Structure}                & \multicolumn{1}{c|}{\textbf{78.06}} & \multicolumn{1}{c|}{71.3}         & \textbf{6.76}                   \\ \hline
\multicolumn{1}{l|}{Det.-Noun   Agr.}                & \multicolumn{1}{c|}{\textbf{97.75}} & \multicolumn{1}{c|}{93.1}         & \textbf{4.65}                   \\ \hline
\multicolumn{1}{l|}{Anaphor   Agr.}                  & \multicolumn{1}{c|}{\textbf{93.61}} & \multicolumn{1}{c|}{89.5}         & \textbf{4.11}                   \\ \hline
\multicolumn{1}{l|}{Ellipsis}                        & \multicolumn{1}{c|}{77.02}          & \multicolumn{1}{c|}{83.8}         & -6.78                           \\ \hline
\multicolumn{1}{l|}{Island   Effects}                & \multicolumn{1}{c|}{45.85}          & \multicolumn{1}{c|}{54.5}         & -8.65                           \\ \hline
\multicolumn{4}{c}{\textbf{BLiMP   Supplement}}                                                                                                                  \\ \hline
\multicolumn{1}{l|}{Sub.   Aux Inversion}            & \multicolumn{1}{c|}{\textbf{77.73}} & \multicolumn{1}{c|}{45.6}         & \textbf{32.13}                  \\ \hline
\multicolumn{1}{l|}{QA  Cong. Easy}                  & \multicolumn{1}{c|}{\textbf{62.5}}  & \multicolumn{1}{c|}{34.4}         & \textbf{28.1}                   \\ \hline
\multicolumn{1}{l|}{Turn   Taking}                   & \multicolumn{1}{c|}{\textbf{62.5}}  & \multicolumn{1}{c|}{46.8}         & \textbf{15.7}                   \\ \hline
\multicolumn{4}{c}{\textbf{GLUE}}                                                                                                                                \\ \hline
\multicolumn{1}{l|}{BoolQ}                           & \multicolumn{1}{c|}{\textbf{65.84}} & \multicolumn{1}{c|}{59.9}         & \textbf{5.94}                   \\ \hline
\multicolumn{1}{l|}{MNLI}                            & \multicolumn{1}{c|}{\textbf{73.73}} & \multicolumn{1}{c|}{68.7}         & \textbf{5.03}                   \\ \hline
\multicolumn{1}{l|}{MNLI-mm}                         & \multicolumn{1}{c|}{74.76}          & \multicolumn{1}{c|}{78}           & -3.24                           \\ \hline
\multicolumn{1}{l|}{QNLI}                            & \multicolumn{1}{c|}{76.86}          & \multicolumn{1}{c|}{82.3}         & -5.44                           \\ \hline
\multicolumn{1}{l|}{RTE}                             & \multicolumn{1}{c|}{45.45}          & \multicolumn{1}{c|}{51.5}         & -6.05                           \\ \hline \hline
\multicolumn{1}{c|}{\textbf{AVG. (overall)}}         & \multicolumn{1}{c|}{\textbf{73.95}} & \multicolumn{1}{c|}{71.75}        & \textbf{2.2}                    \\ \hline \hline

\end{tabular}%
\caption{Selected results of our BabyLM and the RoBERTa (baseline), where the performance of BabyLM improved by at least 3 points (in \textbf{bold}),  or reduced (-) over 3. The metric in this table is all accuracy score. }
\tablabel{result-selected}
\end{table}

We noticed that on the BLiMP benchmark, 5 indicators increased by more than 3 points, namely Filler Gap (+10.52), Subject-Verb Agreement (+8.97), Argument Structure (+6.76), Determiner Noun Agreement (+4.65) and Anaphor Agreement (+4.11), while two tasks dropped by more than 3 points, namely Ellipsis (-6.78) and Island Effects (-8.65). The average performance on this benchmark has also increased by 2.24.

On the GLUE benchmark, BoolQ and MNLI showed an improvement of 5.94 and 5.03, respectively, while MNLI-mm, QNLI, and RTE showed a decrease of  -3.24, -5.44, and -6.05 respectively.


On top of that, we surprisingly find significant improvements in 3 tasks of the BLiMP Supplement benchmark: Subject Aux Inversion (+32.13), QA Congruence Easy (+28.10),  and Turn Taking (+15.70). The average performance on this benchmark improved by 14.85 points.


The overall average performance is increased by 2.2, which shows that our model, pretrained with our reinterpreted small data, already demonstrates a great improvement.

\section{Discussion}
\subsection{Augmented Dataset via CoT Prompting}

We generated our data via the above-mentioned CoT prompting and the GPT-3.5-turbo for nearly 700,~000 lines, we show a case study of a part of the generated data here.

\begin{itemize}
\small
\item[-] \texttt{Paragraph:}
\begin{itemize}
\item[-] \texttt{We have a few topics to cover in this paragraph. Firstly, a possible I.D. has been found in one of Gina's snapshots. Secondly, there is a new technology in development called autostereoscopic 3D that will allow people to watch 3D movies without glasses. This is great news for those who find wearing 3D glasses uncomfortable and causes eye strain. Unfortunately, the narrator regrets not asking Jean for the details about something. Lastly, the police are seen moving down the main street of Atenco, and we are tracking their movements.}
\end{itemize}
\item[-] \texttt{Plan:}
\item[-] 
\begin{itemize}
\item [\texttt{1.}] \texttt{Introduce the topic of the paragraph}
\item [\texttt{2.}] \texttt{Mention the possible I.D. from Gina's snapshots}
\item [\texttt{3.}] \texttt{Talk about the new technology called autostereoscopic 3D}
\item [\texttt{4.}] \texttt{Mention the difficulty of wearing 3D glasses}
\item [\texttt{5.}] \texttt{Mention the regret of not asking Jean for details}
\item [\texttt{6.}] \texttt{Talk about the police and their movement down the main street of Atenco}
\end{itemize}
\item[-] \texttt{Task:}
\item[-] 
\begin{itemize}
\texttt{Text Classification}
\end{itemize}
\item[-] \texttt{Labels:}
\item[-] 
\begin{itemize}
\item [\texttt{1.}] \texttt{I.D. Mentioned}
\item [\texttt{2.}] \texttt{Technology Mentioned}
\item [\texttt{3.}] \texttt{Regret Expressed}
\item [\texttt{4.}] \texttt{Police Mentioned}
\end{itemize}
\end{itemize}



As we can see from the script, the paragraph is an extension of the input sentences sampled from the original dataset, while the plan and labels generated by the language model are the outlines, where the scenes also are the critical information from the generated paragraph. It means that our approach augmented the original data with interpretation, emphasis, and simplification, with which the model is possible to learn about a story with different versions and sizes and finally get a clearer understanding.

\subsection{Performance in QA Congruence Easy} 
We analyzed the most noticeable improvement of the QA Congruence Easy dataset from the BLiMP Supplement benchmark, and dived deep into each case. This dataset consists of 64 single-choice questions with 20 \textit{what}-questions, 25 \textit{who}-questions, and 19 \textit{where}-questions. Each question contains a question mark, and each answer ends with a period. Each question corresponds to 2 candidate answers, and the boundary of the candidate answers is clear, i.e., for the \textit{what}- and \textit{who}-questions, the answers contain an inanimate or an animate, and for the \textit{where}-questions the answer is a location or a noun phrase. Obviously, the answer to the what-questions should be inanimate, like \textit{a car}, the answer to the who-question should be animate, like \textit{a doctor} or person's name \textit{Sarah}, and the answer to the where-question should be location, like \textit{at home}. The model is expected to select the answer that matches the question. For example, a question is ``\textit{Who did you see?}'' and the candidate answers are 1. ``\textit{A doctor}'', 2. ``\textit{A car}'', and it is clear that the answer should be ``\textit{A doctor}''. The final metric for the evaluation is accuracy.

\subsubsection{Influence of the 3 Types of Questions}

In these three kinds of questions, our model is better at answering the \textit{what}-questions, where the accuracy is 75. Besides, it obtains an accuracy of 64 for the \textit{who}-questions, and 47 for the \textit{where}-questions. 

\subsubsection{Influence of the 2 Types of Answers}

We also note that there are two forms of the answers:

\begin{itemize}
  \item [1)] 
   \textit{sentence}, where the answer is a complete sentence that includes at least the verb, e.g. ``\textit{I sent the package to europe}''; 
  \item [2)]
  \textit{fragment}, where the answer is a single word or a simple phrase, and does not include the verb, e.g. ``\textit{a car}''. 
\end{itemize}

The form of the two candidates' answers to each question is consistent, i.e., both candidates' answers are either sentences or fragments. The dataset contains 27 question-answer pairs in the form of sentences (42\%) and 37 cases in fragments (57\%). We also counted the accuracy on the above two forms, where the accuracy is 77.78 for sentences and 51.35 for fragments. Additionally, we also counted the accuracy with the different forms of the three questions i.e. \textit{what}-, \textit{who}-, and \textit{where}-questions. The accuracy of the sentence labels on the \textit{what}-questions is 80, while the fragment is 70. The accuracy on the who-question with sentence answers was 71 and 61 with fragment answers. On \textit{where}-questions, the tasks with sentence answers obtained an accuracy of 80, however, it was only 11 with the fragment answers. Thus we can observe that our model is better at deciding with complete answers rather than fragments.

\subsubsection{Influence of the 3 Types of Dialogues}

Besides, we also notice that there are three types of dialogues for each question, 
\begin{itemize}
  \item [1)] 
  \textit{direct} dialogues, where the question is started by a question word directly and the answer is direct with the answer, e.g., question: ``\textit{What did you get?}'', candidate answers: ``\textit{I got a chair}'',  ``\textit{I got a doctor}'';        
  \item [2)]
  \textit{A-B} dialogues, where the letters \textit{A} and \textit{B} are used as names for both sides of the conversation before proposing the question and the candidate answers respectively, e.g. question ``\textit{A}: What did you sell?'', candidate answers: ``\textit{B}: A chair.'', ``\textit{B}: A doctor.'';  
  \item [3)]
  \textit{David-Sarah} dialogues, the person's name \textit{David} is used as the questioner's name before the question, and \textit{Sarah} is used as the answerer's name before the answer.
\end{itemize}

The dataset comprises 21 direct dialogues (32\%), 22 \textit{A-B} dialogues (34\%), and 21 \textit{David-Sarah} dialogues (32\%), with the model’s accuracy consistently ranging between 61-63\% across these types.

We then explored the proportionality between these three forms of dialogue and the three kinds of questions. Of the 20 \textit{what}-questions, 7 are written in \textit{direct} dialogues, 6 are in \textit{A-B} dialogues, and 7 are \textit{David-Sarah} dialogues. we notice a difference in the accuracy, where the accuracy with \textit{direct} dialogues is 100, the \textit{A-B} dialogues have an accuracy of 83, and the David-Sarah dialogues reached only 45. 

Of the 25 \textit{who}-questions, 8 \textit{direct} dialogues obtained an accuracy only of 25, while 7 \textit{A-B} dialogues gained 85 accuracy and the accuracy of the 10 \textit{David-Sarah} dialogues is 80. Out of the 19 \textit{where}-questions, the accuracy of the 6 direct dialogues is 66\%, 33\% of \textit{A-B} dialogues are correct, and the accuracy of the 4 \textit{David-Sarah} dialogues is 50\%.

From the above results, we can see that our model is good at selecting answers from \textit{direct} and \textit{A-B} dialogues on the \textit{what}-questions. In contrast, for the \textit{who}-questions, our model is good at selecting animates from the \textit{David-Sarah} dialogues and the \textit{A-B} dialogues, but not good at selecting the animate from the direct dialogues. It might be positively affected by the presence of the person's name. In the \textit{where}-questions, the form of dialogues has a more limited effect on the performance. 

\subsection{Performance in QA Congruence Tricky} 

We compared the performance on the QA Congruence Tricky dataset, on which we have a very similar performance (35) to the baseline model. It contains 165 tricky questions including \textit{who}-, \textit{where}-, \textit{when}-, \textit{why}-, and \textit{how many}-questions, where the proportions of the \textit{who}- and the \textit{where}-questions are 15\% and 16\% respectively. The accuracy of the \textit{who}- and \textit{where}-questions are only 37 and 30 respectively, differ from the accuracies in the QA Congruence Easy dataset. 

We also notice that, in this dataset, our model is better at selecting fragment answers rather than answers in the form of sentences, where the accuracy with fragments is 62, while the accuracy of the sentences is only 10. On both \textit{who}- and \textit{where}-questions, our model is better at finding the answer in the \textit{David-Sarah} dialogues (55 and 45 respectively in accuracy), and the accuracies of both questions in the other two dialogue forms are under 30.  
Similar to the fact shown in the easy dataset, the presence of people's names probably provides a sign to the animate and thus influences the performance, especially on the \textit{who}-questions.  

We analyzed the questions-candidate answers pairs from the tricky dataset, where both the questions and the candidate answers are generally shorter, e.g., the question is ``\textit{Who ate?}'', and the candidate answers are ``\textit{A teacher ate}.'', and ``\textit{Pasta ate.}'', where the question only contains the \textit{wh}-word, a verb, and a question mark, and the candidate answers contain only a subjective and a verb. The answers in the form of fragments are even shorter, e.g. to a question ``\textit{Who cooked?}'', the candidate answers are ``\textit{Sarah}'', and ``\textit{A sandwich}''. 

Besides the questions being more varied and complex, this dataset is more tricky, because the context is short. The candidate answers written in sentences are generally very similar to the fragments with only an additional verb, where the verb has been mentioned in the questions, which means the form of sentence possibly doesn't provide additional information, but may confuse the model to understand the answers.


\section{Conclusion}
In this work, we proposed the CoThought pipeline for training a BabyLM at a small scale, combining the LLMs' productivity with the concept of a child’s cognitive learning ability. We let the raw training data for the BabyLM be reformulated by the LLM's CoT prompting (i.e. let the teacher think) and then train a BabyLM in a pretraining fashion based on the newly structured data (i.e. let the child co-think and learn). We compare the performance results of our BabyLM to another vanilla pretrained LM RoBERTa and demonstrate that our model achieves higher performance in many tasks including linguistic, question and answer, especially congruence tasks. This suggests that data processed by LLMs based on their contextual reasoning is more natural and efficient in the learning process, just as text revised by experienced teachers in the school is more suitable for students to learn and understand. And when we use data restructured by LLMs, even in the case of small data volume, the model is able to achieve the effect of a model trained from a large amount of data, or to be even better.

\section*{Limitations}
One limitation of our work is the exclusive use of a specific LLM for data generation. It would be insightful to explore how performance varies when using different LLMs to generate the pre-training data. Different LLMs may introduce variability and diversity in the generated data, which could influence the effectiveness of the pre-training process. This aspect, while not explored in our current work, presents a promising avenue for future research to understand the impact of various LLMs on data generation and subsequent model performance.

Another limitation of our work is that our primary focus is on data generation, leaving potential improvements or optimizations in this domain unexplored.

Additionally, our model training exclusively utilized the RoBERTa architecture. Other architectures, including causal language models and various transformer variants, also showed potential research value. Therefore, exploring our approach across a broader range of architectures and identifying pretraining methods most compatible with our generated data remains an important area for future research.

By acknowledging these limitations, we hope to spur further research in this area, encouraging the exploration of data generation techniques, model architectures, and extended data methods in the context of small-scale language modeling.

\section*{Ethics Statement}
This research was conducted in accordance with the ACM Code of Ethics. 
The datasets that we use are publicly available \citep{warstadt-et-al-2023-babylm}. 
We report only aggregated results in the main paper.
We have not intended or do not intend to share any Personally Identifiable Data with this paper.

\section*{Acknowledgements}
We thank the anonymous reviewers and the organizing committee for their efforts and helpful advice. E.N. was supported by MCML and CSC.

\bibliography{custom}
\bibliographystyle{acl_natbib}


\appendix

\section{Code and Model}
The code for data processing and model training is available at: \url{https://github.com/oooranz/Baby-CoThought}.

Our BabyLM is available at: \url{https://huggingface.co/yaanhaan/Baby-CoThought}.

\section{Pretraining Data Statistics}
The generated dataset for LM pretraining is available at: \url{https://huggingface.co/datasets/yaanhaan/Baby-CoThought-Data}. 

We present a statistical analysis of the generated dataset. Given that our task revolves around creative NLU example generation, the dataset inherently encompasses a wide variety of tasks. This diversity is reflective of the creative nature of the task, allowing for a richer and more comprehensive pretraining process. Each example in the dataset includes an NLU example and its corresponding reason. 

We plot the task distribution of the pretraining dataset in \figref{dist}. Tasks that appeared only once in the dataset are categorized as others.

\begin{figure}[ht]
\centering
\includegraphics[width=0.47\textwidth]{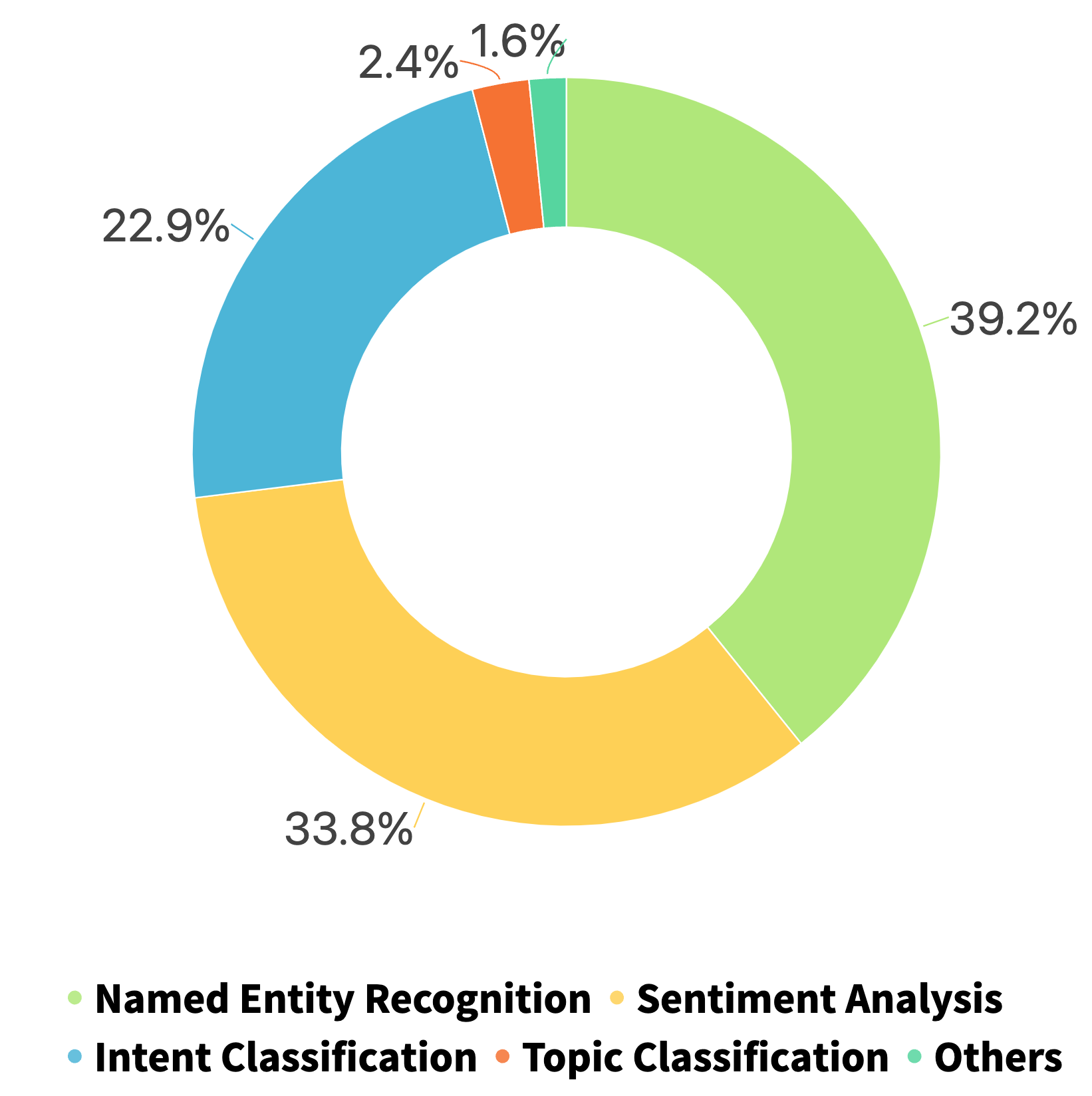}
\caption{The distribution of the different NLU task examples in the pretraining dataset.}
\figlabel{dist}
\end{figure}

The average number of words in the paragraphs across all examples in the dataset is approximately 115.25 words.

\section{Hyperparameter}
\seclabel{hyperparameter}
We followed the instruction\footnote{\url{https://huggingface.co/blog/how-to-train}} and trained the tokenizers separately for the original dataset and our enhanced dataset via the \texttt{ByteLevelBPETokenizer} library with the hyperparameters shown in \tabref{hyperparameter_tokenizer}. Other hyperparameters were set to default and can be found in the document 
\footnote{\url{https://github.com/huggingface/tokenizers/blob/main/bindings/python/py_src/tokenizers/implementations/byte_level_bpe.py}}.

\begin{table}[ht]
\small
\centering
\renewcommand\arraystretch{1.35}
\begin{tabular}{lr}
\hline \hline
\textbf{Hyperparameter} & \textbf{Value}                                                              \\ \hline
vocab\_size             & 52000\\
min\_frequency          & 2\\
special\_tokens         & \texttt{<s>}, \texttt{<pad>}, \texttt{</s>}, \texttt{<unk>},\texttt{<mask>} 
\\ 
\hline \hline
\end{tabular}
\caption{Hyperparameters used for tokenizers}
\tablabel{hyperparameter_tokenizer}
\end{table}

Besides, we report our hyperparameters during the pretraining of our RoBERTa models in \tabref{hyperparameter}. We used the default settings from the \texttt{RobertaConfig} library. More default values and technical details can be found in the documents 3111\footnote{\url{https://huggingface.co/docs/transformers/model\_doc/roberta\#transformers.RobertaConfig}}.

\begin{table}[ht]
\small
\centering
\renewcommand\arraystretch{1.35}
\begin{tabular}{lr}
\hline \hline
\textbf{Hyperparameter}                  & \textbf{Value}    \\ 
\hline
attention\_probs\_dropout\_prob & 0.1      \\ 
bos\_token\_id                  & 0        \\ 
classifier\_dropout             & null     \\ 
eos\_token\_id                  & 2        \\ 
hidden\_act                     & gelu     \\ 
hidden\_dropout\_prob           & 0.1      \\ 
hidden\_size                    & 768      \\ 
initializer\_range              & 0.02     \\ 
intermediate\_size              & 3072     \\ 
layer\_norm\_eps                & 1.00E-12 \\ 
max\_position\_embeddings       & 512      \\ 
model\_type                     & roberta  \\ 
num\_attention\_heads           & 12       \\ 
num\_hidden\_layers             & 12       \\ 
pad\_token\_id                  & 1        \\ 
position\_embedding\_type       & absolute \\ 
torch\_dtype                    & float32  \\ 
transformers\_version           & 4.17.0   \\ 
type\_vocab\_size               & 1        \\ 
use\_cache                      & TRUE     \\ 
vocab\_size                     & 52000    \\ 
\hline \hline
\end{tabular}%
\caption{Hyperparameters used for pretraining}
\tablabel{hyperparameter}
\end{table}

Additionally, the evaluation process was done automatically via the evaluation tool provided by the organizer, without changing the hyperparameters, which can be found on the webpage \footnote{\url{https://github.com/babylm/evaluation-pipeline\#hyperparameters}}.

\section{Full Results}
\seclabel{full_results}

We used 4 benchmarks:

\begin{itemize}
  \item [1)] 
   Benchmark of Linguistic Minimal Pairs (BLiMP) \cite{warstadt-etal-2020-blimp-benchmark}, including Anaphor Agreement, Argument Structure, Binding, Control Raising, Determiner Noun Agreement, Ellipsis, Filler Gap, Irregular Forms, Island Effects, NPI Licensing, Quantifiers, and Subject Verb Agreement; 
  \item [2)]
   BLiMP Supplement\footnote{\url{https://github.com/babylm/evaluation-pipeline/blob/main/filter\_data.zip}}, including Hypernym, QA Congruence Easy, QA Congruence Tricky, Subject Aux Inversion, and Turn Taking; 
  \item [3)]
    General Language Understanding Evaluation (GLUE) \cite{GLUE}, including CoLA \cite{COLA}, SST-2 \cite{SST-2}, MRPC (F1) \cite{MRPC}, QQP\footnote{\url{https://quoradata.quora.com/First-Quora-Dataset-Release-Question-Pairs}} (F1), MNLI \cite{MNLI}, MNLI-mm, QNLI \cite{NLI}, RTE \cite{RTE-1, RTE-2, RTE-3, RTE-4}, BoolQ \cite{BoolQ}, MultiRC \cite{MultiRC} and WSC \cite{WSC};
  \item [4)]
    Mixed Signals Generalization Set (MSGS)  \cite{MSGS}, including Control Raising Control (CR Control), Lexical Content The Control (LC Control), Main Verb Control (MV Control), Relative Position Control (RP Control), Syntactic Category Control (SC Control), Control Raising Lexical Content The (CR LC), Control Raising Relative Token Position (CR RTP), Main Verb Lexical Content The (MV LC), Main Verb Relative Token Position (MV RTP), Syntactic Category Lexical Content The (SC LC), Syntactic Category Relative Position (SC RP). 
\end{itemize}
to process our evaluation. 

The organizer provided three baseline models, including OPT-125M\footnote{\url{https://huggingface.co/facebook/opt-125m}} , RoBERTa-base\footnote{\url{https://huggingface.co/roberta-base}} , and T5-base\footnote{\url{https://huggingface.co/t5-base}}. 
We show our full results in \tabref{full-result}.

\begin{table*}[ht]
\small
\centering
\renewcommand\arraystretch{1.25}
\setlength\tabcolsep{8pt}
\begin{tabular}{lcccccc}

\hline \hline
\multicolumn{1}{c|}{\multirow{2}{*}{\textbf{Tasks}}} & \multicolumn{4}{c|}{\textbf{Models}}                                                                                                   & \multicolumn{2}{c}{\textbf{Difference}}                \\ \cline{2-7} 
\multicolumn{1}{c|}{}                                  & \multicolumn{1}{c|}{Ours}           & \multicolumn{1}{c|}{OPT-125m} & \multicolumn{1}{c|}{RoBERTa-base} & \multicolumn{1}{c|}{T5-base} & \multicolumn{1}{c|}{in abs}         & in rel.          \\ \hline
\multicolumn{7}{c}{\textbf{BLiMP}}                                                                                                                                                                                                                       \\ \hline
\multicolumn{1}{l|}{Anaphor Agreement}                 & \multicolumn{1}{c|}{93.61}          & \multicolumn{1}{c|}{\textbf{94.90}}    & \multicolumn{1}{c|}{89.50}        & \multicolumn{1}{c|}{66.70}   & \multicolumn{1}{c|}{4.11}  & 4.59\%  \\ \hline
\multicolumn{1}{l|}{Argument Structure}                & \multicolumn{1}{c|}{\textbf{78.06}}          & \multicolumn{1}{c|}{73.80}    & \multicolumn{1}{c|}{71.30}        & \multicolumn{1}{c|}{61.20}   & \multicolumn{1}{c|}{6.76}  & 9.48\%  \\ \hline
\multicolumn{1}{l|}{Binding}                           & \multicolumn{1}{c|}{72.84}          & \multicolumn{1}{c|}{\textbf{73.80}}    & \multicolumn{1}{c|}{71.00}        & \multicolumn{1}{c|}{59.40}   & \multicolumn{1}{c|}{1.84}  & 2.59\%  \\ \hline
\multicolumn{1}{l|}{Control Raising}                   & \multicolumn{1}{c|}{69.55}          & \multicolumn{1}{c|}{\textbf{72.20}}    & \multicolumn{1}{c|}{67.10}        & \multicolumn{1}{c|}{59.80}   & \multicolumn{1}{c|}{2.45}  & 3.65\%  \\ \hline
\multicolumn{1}{l|}{Determiner Noun Agreement}         & \multicolumn{1}{c|}{\textbf{97.75}}          & \multicolumn{1}{c|}{93.10}    & \multicolumn{1}{c|}{93.10}        & \multicolumn{1}{c|}{53.80}   & \multicolumn{1}{c|}{4.65}  & 4.99\%  \\ \hline
\multicolumn{1}{l|}{Ellipsis}                          & \multicolumn{1}{c|}{77.02}          & \multicolumn{1}{c|}{80.50}    & \multicolumn{1}{c|}{\textbf{83.80}}        & \multicolumn{1}{c|}{49.10}   & \multicolumn{1}{c|}{-6.78} & -8.09\%          \\ \hline
\multicolumn{1}{l|}{Filler Gap}                        & \multicolumn{1}{c|}{\textbf{78.52}}          & \multicolumn{1}{c|}{73.60}    & \multicolumn{1}{c|}{68.00}        & \multicolumn{1}{c|}{70.00}   & \multicolumn{1}{c|}{10.52} & 15.47\% \\ \hline
\multicolumn{1}{l|}{Irregular Forms}                   & \multicolumn{1}{c|}{\textbf{91.25}}          & \multicolumn{1}{c|}{80.80}    & \multicolumn{1}{c|}{89.60}        & \multicolumn{1}{c|}{75.50}   & \multicolumn{1}{c|}{1.65}  & 1.84\%  \\ \hline
\multicolumn{1}{l|}{Island Effects}                    & \multicolumn{1}{c|}{45.85}          & \multicolumn{1}{c|}{\textbf{57.80}}    & \multicolumn{1}{c|}{54.50}        & \multicolumn{1}{c|}{43.60}   & \multicolumn{1}{c|}{-8.65} & -15.87\%         \\ \hline
\multicolumn{1}{l|}{NPI Licensing}                     & \multicolumn{1}{c|}{\textbf{67.35}}          & \multicolumn{1}{c|}{51.60}    & \multicolumn{1}{c|}{66.30}        & \multicolumn{1}{c|}{45.60}   & \multicolumn{1}{c|}{1.05}  & 1.58\%  \\ \hline
\multicolumn{1}{l|}{Quantifiers}                       & \multicolumn{1}{c|}{70.58}          & \multicolumn{1}{c|}{\textbf{74.50}}    & \multicolumn{1}{c|}{70.30}        & \multicolumn{1}{c|}{34.20}   & \multicolumn{1}{c|}{0.28}  & 0.40\%  \\ \hline
\multicolumn{1}{l|}{Subject Verb Agreement}            & \multicolumn{1}{c|}{\textbf{85.17}}          & \multicolumn{1}{c|}{77.30}    & \multicolumn{1}{c|}{76.20}        & \multicolumn{1}{c|}{53.20}   & \multicolumn{1}{c|}{8.97}  & 11.77\% \\ \hline
\multicolumn{7}{c}{\textbf{BLiMP Supplement}}                                                                                                                                                                                                            \\ \hline
\multicolumn{1}{l|}{Hypernym}                          & \multicolumn{1}{c|}{49.07}          & \multicolumn{1}{c|}{46.30}    & \multicolumn{1}{c|}{50.80}        & \multicolumn{1}{c|}{\textbf{51.10}}   & \multicolumn{1}{c|}{-1.73} & -3.41\%          \\ \hline
\multicolumn{1}{l|}{QA Congruence Easy}                & \multicolumn{1}{c|}{62.50}          & \multicolumn{1}{c|}{\textbf{76.50}}    & \multicolumn{1}{c|}{34.40}        & \multicolumn{1}{c|}{45.30}   & \multicolumn{1}{c|}{28.10} & 81.69\% \\ \hline
\multicolumn{1}{l|}{QA Congruence Tricky}              & \multicolumn{1}{c|}{34.55}          & \multicolumn{1}{c|}{\textbf{47.90}}    & \multicolumn{1}{c|}{34.50}        & \multicolumn{1}{c|}{25.50}   & \multicolumn{1}{c|}{0.05}  & 0.14\%  \\ \hline
\multicolumn{1}{l|}{Subject Aux Inversion}             & \multicolumn{1}{c|}{\textbf{77.73}}          & \multicolumn{1}{c|}{85.30}    & \multicolumn{1}{c|}{45.60}        & \multicolumn{1}{c|}{69.20}   & \multicolumn{1}{c|}{32.13} & 70.46\% \\ \hline
\multicolumn{1}{l|}{Turn Taking}                       & \multicolumn{1}{c|}{62.50}          & \multicolumn{1}{c|}{\textbf{82.90}}    & \multicolumn{1}{c|}{46.80}        & \multicolumn{1}{c|}{48.90}   & \multicolumn{1}{c|}{15.70} & 33.55\% \\ \hline
\multicolumn{7}{c}{\textbf{GLUE}}                                                                                                                                                                                                                        \\ \hline
\multicolumn{1}{l|}{CoLA}                              & \multicolumn{1}{c|}{74.09}          & \multicolumn{1}{c|}{73.70}    & \multicolumn{1}{c|}{75.90}        & \multicolumn{1}{c|}{\textbf{76.30}}   & \multicolumn{1}{c|}{-1.81} & -2.38\%          \\ \hline
\multicolumn{1}{l|}{SST-2}                             & \multicolumn{1}{c|}{\textbf{88.78}}          & \multicolumn{1}{c|}{86.60}    & \multicolumn{1}{c|}{88.60}        & \multicolumn{1}{c|}{88.00}   & \multicolumn{1}{c|}{0.18}  & 0.20\%  \\ \hline
\multicolumn{1}{l|}{MRPC (F1)}                         & \multicolumn{1}{c|}{80.45}          & \multicolumn{1}{c|}{82.10}    & \multicolumn{1}{c|}{80.50}        & \multicolumn{1}{c|}{\textbf{85.90}}   & \multicolumn{1}{c|}{-0.05} & -0.06\%          \\ \hline
\multicolumn{1}{l|}{QQP (F1)}                          & \multicolumn{1}{c|}{\textbf{81.20}}          & \multicolumn{1}{c|}{77.80}    & \multicolumn{1}{c|}{78.50}        & \multicolumn{1}{c|}{79.70}   & \multicolumn{1}{c|}{2.70}  & 3.44\%  \\ \hline
\multicolumn{1}{l|}{MNLI}                              & \multicolumn{1}{c|}{\textbf{73.73}}          & \multicolumn{1}{c|}{70.10}    & \multicolumn{1}{c|}{68.70}        & \multicolumn{1}{c|}{71.50}   & \multicolumn{1}{c|}{5.03}  & 7.32\%  \\ \hline
\multicolumn{1}{l|}{MNLI-mm}                           & \multicolumn{1}{c|}{74.76}          & \multicolumn{1}{c|}{71.90}    & \multicolumn{1}{c|}{\textbf{78.00}}        & \multicolumn{1}{c|}{74.00}   & \multicolumn{1}{c|}{-3.24} & -4.15\%          \\ \hline
\multicolumn{1}{l|}{QNLI}                              & \multicolumn{1}{c|}{76.86}          & \multicolumn{1}{c|}{80.10}    & \multicolumn{1}{c|}{82.30}        & \multicolumn{1}{c|}{\textbf{83.10}}   & \multicolumn{1}{c|}{-5.44} & -6.61\%          \\ \hline
\multicolumn{1}{l|}{RTE}                               & \multicolumn{1}{c|}{45.45}          & \multicolumn{1}{c|}{\textbf{67.70}}    & \multicolumn{1}{c|}{51.50}        & \multicolumn{1}{c|}{60.60}   & \multicolumn{1}{c|}{-6.05} & -11.74\%         \\ \hline
\multicolumn{1}{l|}{BoolQ}                             & \multicolumn{1}{c|}{65.84}          & \multicolumn{1}{c|}{66.00}    & \multicolumn{1}{c|}{59.90}        & \multicolumn{1}{c|}{\textbf{69.00}}   & \multicolumn{1}{c|}{5.94}  & 9.91\%  \\ \hline
\multicolumn{1}{l|}{MultiRC}                           & \multicolumn{1}{c|}{62.21}          & \multicolumn{1}{c|}{61.10}    & \multicolumn{1}{c|}{61.30}        & \multicolumn{1}{c|}{\textbf{62.40}}   & \multicolumn{1}{c|}{0.91}  & 1.49\%  \\ \hline
\multicolumn{1}{l|}{WSC}                               & \multicolumn{1}{c|}{\textbf{61.45}}          & \multicolumn{1}{c|}{59.00}    & \multicolumn{1}{c|}{61.40}        & \multicolumn{1}{c|}{60.20}   & \multicolumn{1}{c|}{0.05}  & 0.07\%  \\ \hline
\multicolumn{7}{c}{\textbf{MSGS}}                                                                                                                                                                                                                        \\ \hline
\multicolumn{1}{l|}{CR (Control)}                      & \multicolumn{1}{c|}{83.96}          & \multicolumn{1}{c|}{\textbf{97.20}}    & \multicolumn{1}{c|}{93.00}        & \multicolumn{1}{c|}{95.10}   & \multicolumn{1}{c|}{-9.04} & -9.72\%          \\ \hline
\multicolumn{1}{l|}{LC (Control)}                      & \multicolumn{1}{c|}{94.49}          & \multicolumn{1}{c|}{82.60}    & \multicolumn{1}{c|}{\textbf{100.00}}       & \multicolumn{1}{c|}{\textbf{100.00}}  & \multicolumn{1}{c|}{-5.51} & -5.51\%          \\ \hline
\multicolumn{1}{l|}{MV (Control)}                      & \multicolumn{1}{c|}{99.98}          & \multicolumn{1}{c|}{\textbf{100.00}}   & \multicolumn{1}{c|}{\textbf{100.00}}       & \multicolumn{1}{c|}{\textbf{100.00}}  & \multicolumn{1}{c|}{-0.02} & -0.02\%          \\ \hline
\multicolumn{1}{l|}{RP (Control)}                      & \multicolumn{1}{c|}{\textbf{100.00}}         & \multicolumn{1}{c|}{99.80}    & \multicolumn{1}{c|}{\textbf{100.00}}       & \multicolumn{1}{c|}{99.80}   & \multicolumn{1}{c|}{0.00}  & 0.00\%           \\ \hline
\multicolumn{1}{l|}{SC (Control)}                      & \multicolumn{1}{c|}{88.44}          & \multicolumn{1}{c|}{88.10}    & \multicolumn{1}{c|}{\textbf{89.00}}        & \multicolumn{1}{c|}{88.70}   & \multicolumn{1}{c|}{-0.56} & -0.62\%          \\ \hline
\multicolumn{1}{l|}{CR LC}                             & \multicolumn{1}{c|}{67.07}          & \multicolumn{1}{c|}{75.30}    & \multicolumn{1}{c|}{68.30}        & \multicolumn{1}{c|}{\textbf{76.70}}   & \multicolumn{1}{c|}{-1.23} & -1.80\%          \\ \hline
\multicolumn{1}{l|}{CR RTP}                            & \multicolumn{1}{c|}{\textbf{70.71}}          & \multicolumn{1}{c|}{67.10}    & \multicolumn{1}{c|}{66.80}        & \multicolumn{1}{c|}{69.40}   & \multicolumn{1}{c|}{3.91}  & 5.86\%  \\ \hline
\multicolumn{1}{l|}{MV LC}                             & \multicolumn{1}{c|}{66.61}          & \multicolumn{1}{c|}{66.30}    & \multicolumn{1}{c|}{66.60}        & \multicolumn{1}{c|}{\textbf{67.00}}   & \multicolumn{1}{c|}{0.01}  & 0.01\%           \\ \hline
\multicolumn{1}{l|}{MV RTP}                            & \multicolumn{1}{c|}{67.59}          & \multicolumn{1}{c|}{66.80}    & \multicolumn{1}{c|}{\textbf{80.20}}        & \multicolumn{1}{c|}{67.70}   & \multicolumn{1}{c|}{-12.61}& -15.72\%         \\ \hline
\multicolumn{1}{l|}{SC LC}                             & \multicolumn{1}{c|}{75.47}          & \multicolumn{1}{c|}{\textbf{84.80}}    & \multicolumn{1}{c|}{67.40}        & \multicolumn{1}{c|}{72.70}   & \multicolumn{1}{c|}{8.07}  & 11.98\% \\ \hline
\multicolumn{1}{l|}{SC RP}                             & \multicolumn{1}{c|}{\textbf{70.90}}          & \multicolumn{1}{c|}{62.00}    & \multicolumn{1}{c|}{67.40}        & \multicolumn{1}{c|}{68.00}   & \multicolumn{1}{c|}{3.50}  & 5.19\%  \\ \hline \hline

\end{tabular}
\caption{Full results, with difference of our BabyLM over RoBERTa-base (baseline). Metric of MRPC and QQP from GLUE is $F_1$, in other tasks the metric is accuracy. 
The best results of the four models are marked in \textbf{bold}.
}
\tablabel{full-result}
\end{table*}

\end{document}